\documentclass[10pt,twocolumn]{article}
\usepackage{graphicx}
\usepackage{amsmath}  
\usepackage{hyperref} 
\usepackage[affil-it]{authblk} 
\usepackage{amssymb}
\usepackage{booktabs}
\usepackage{comment}
\usepackage{multirow}
\usepackage{makecell}
\usepackage{cleveref}
\usepackage{float}
\usepackage[utf8]{inputenc}
\usepackage[margin=0.75in]{geometry}

\renewcommand{\abstractname}{\large\textbf{Abstract}}
\renewenvironment{abstract}
  {\center\Large\textbf{\abstractname}\par\normalsize\par\noindent}
  {\vspace{0.5cm}}
\makeatletter
\renewcommand\title[1]{\gdef\@title{\Large\bfseries #1}} 
\makeatother

\title{SRA: A Novel Method to Improve Feature Embedding in Self-supervised Learning for Histopathological Images}

\begin{document}

\author{
    \upshape{Hamid Manoochehri}\textsuperscript{*1,2} \quad {Bodong Zhang}\textsuperscript{*\dag1,2} \quad {Beatrice S. Knudsen}\textsuperscript{3} \quad {Tolga Tasdizen}\textsuperscript{1,2}\\\vspace{5pt}
    {\fontsize{10.5pt}{12pt}\selectfont \textsuperscript{1}Department of Electrical and Computer Engineering, University of Utah, Salt Lake City, UT, USA\\ \textsuperscript{2}Scientific Computing and Imaging Institute, University of Utah, Salt Lake City, UT, USA\\ \textsuperscript{3}Department of Pathology, University of Utah, Salt Lake City, UT, USA
    \\\upshape\texttt{hamid.manoochehri@utah.edu}, \texttt{bodong.zhang@utah.edu}, \\\upshape\texttt{beatrice.knudsen@pathology.utah.edu}, \texttt{tolga.tasdizen@utah.edu}
    }}

\date{}

\twocolumn[
  \begin{@twocolumnfalse}
    \maketitle
    \begin{abstract}
    Self-supervised learning has become a cornerstone in various areas, particularly histopathological image analysis. Image augmentation plays a crucial role in self-supervised learning, as it generates variations in image samples. However, traditional image augmentation techniques often overlook the unique characteristics of histopathological images. In this paper, we propose a new histopathology-specific image augmentation method called stain reconstruction augmentation (SRA). We integrate our SRA with MoCo v3, a leading model in self-supervised contrastive learning, along with our additional contrastive loss terms, and call the new model SRA-MoCo v3. We demonstrate that our SRA-MoCo v3 always outperforms the standard MoCo v3 across various downstream tasks and achieves comparable or superior performance to other foundation models pre-trained on significantly larger histopathology datasets. 
    \end{abstract}
    \vspace{0.5cm}
  \end{@twocolumnfalse}
]

\renewcommand{\thefootnote}{\fnsymbol{footnote}}
\footnotetext{*The first two authors contributed equally to this work.}
\footnotetext{\dag Corresponding author. Email: \nolinkurl{bodong.zhang@utah.edu}}

\section{Introduction}
\label{sec:tileexamples_intro}
Deep learning serves as an invaluable tool for medical diagnosis, including cancer detection and grading. Developing robust models for these purposes is critically important and has gained widespread popularity. Possessing such accurate classification models allows medical doctors to expedite cancer diagnoses and make appropriate prognostic decisions.

Self-supervised learning (SSL) \cite{Jing2021-ix-selfsup-survey,Gui2024-eq-selfsup-survey} has become a mainstream technique in deep learning. One of its key advantages is that it eliminates the need for labeled data during pre-training by generating labels directly from the data itself. 
Another important advantage of self-supervised learning is its capacity to enable transfer learning through pre-trained models. By leveraging large-scale datasets and extensive computational resources for pre-training, self-supervised learning models capture rich and generalizable feature representations. Once trained, these models can be shared and deployed by researchers or practitioners who lack access to large datasets or significant computational power.
Contrastive learning \cite{Jaiswal2020-ha-contrastive-survey} has become particularly successful in self-supervised pre-training, as it aims to maximize the similarity between different augmentations of the same data points (positive pairs) and minimize the similarity between views of different data points (negative pairs). This encourages the model to learn robust and discriminative feature representations, which can perform well across a variety of downstream tasks.

Data augmentation is crucial for self-supervised contrastive learning, as it introduces variability for each data point and helps the model generalize better to unseen data \cite{Wagner2022-mt}. Standard image augmentation techniques apply general color, morphological, or geometric transformations, or a combination of these, such as rotation, cropping, resizing, and color jittering. However, these methods do not account for the specific characteristics of histopathological images \cite{Tellez2019-co}.

\begin{figure}[ht]
    \centering
    \includegraphics[width=1.0\linewidth]{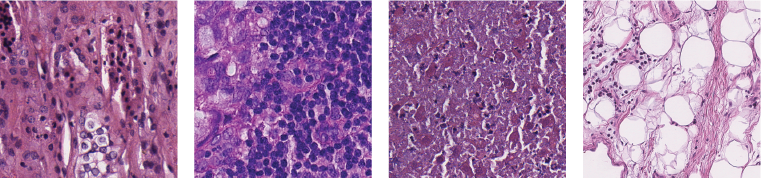}
    \caption{Patch examples from various datasets. }
    \label{fig:tile_examples}
\end{figure}

Hematoxylin \& Eosin (H\&E) stained histopathological image is the dominant image type in histopathology. Hematoxylin stains the cell nuclei a deep blue or purple, while Eosin stains the cytoplasm and extracellular matrix pink, allowing for clear differentiation between different cellular and tissue components. The variations of stains across images are shown in \cref{fig:tile_examples}.

In this paper, we propose a novel histopathology-specific image augmentation method called stain reconstruction augmentation (SRA) and apply it to H\&E stained histopathological images. We first performed stain separation \cite{Ruifrok2001-lm} to decompose RGB images into Hematoxylin stain channel images and Eosin stain channel images in Optical Density (OD) space. Instead of multiplying a preset random scalar to each channel for augmentation \cite{tellez2018,Tellez2018-wj,Chang2021-th-mixup}, where the augmented images are statistically affected by the intensities of original stain channel images, we first normalize each single stain channel image by dividing it by its max intensity, then multiplying the random scalar we predefined to augment each stain channel image. By this way, the intensities of augmented stain channel images are not affected by the max intensity of original stain channel images. Additionally, when we reconstruct augmented images from OD space back to RGB space, we define a probability for excluding one stain channel and only allow the other channel to remain to provide stronger augmentations. We integrate our SRA into MoCo v3 \cite{Chen2021-gz-mocov3}, one of the state-of-the-art self-supervised contrastive learning models, as well as our additional contrastive loss terms, to form our new model SRA-MoCo v3. 

The main contributions of our paper are as below:  
\begin{itemize}
    \item We propose a novel histopathological image augmentation method called stain reconstruction augmentation (SRA) and integrate it into MoCo v3 for better self-supervised contrastive learning on digital histopathological images.
    \item Original MoCo v3's loss function only calculates contrastive loss between feature sets obtained from different augmentations by different encoders (momentum and query encoders). To further stabilize the learned features after SRA, we introduce additional loss terms calculating the contrastive loss between features obtained from different augmentations of the images by the same encoder.
    \item We conducted experiments on multiple publicly available datasets. We demonstrate that our proposed SRA-MoCo v3 always outperforms MoCo v3 on various downstream tasks. We also achieved comparable or superior performance to other foundation models pre-trained on significantly larger histopathology datasets.
    \item We made the code publicly available at \hyperlink{https://github.com/hamidmanoochehri/Paper_SRA}{github.com/hamidmanoochehri/Paper\_SRA}
\end{itemize}

\begin{figure*}[ht]
{\small{
    \centering
    \includegraphics[width=1.0\linewidth]{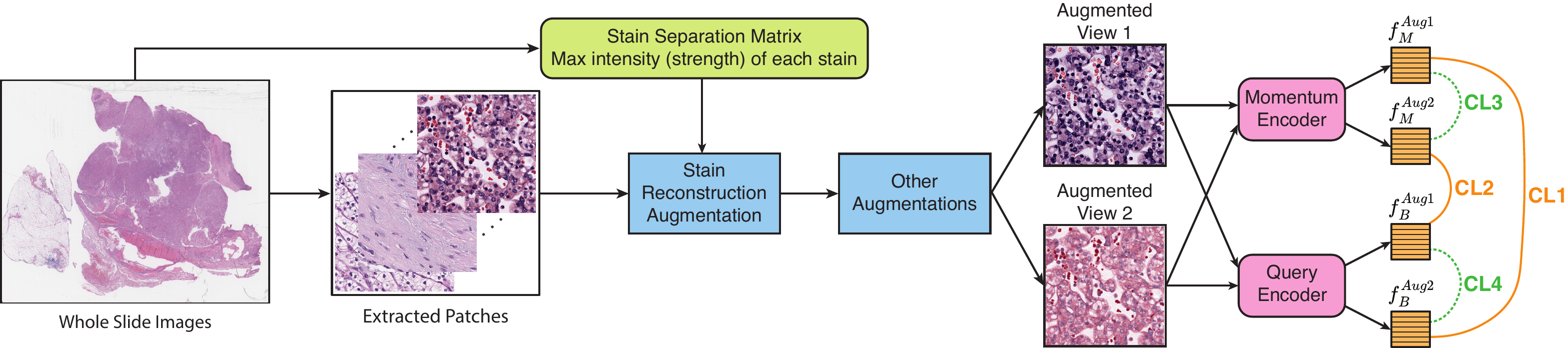}
    \caption{The pipeline of our SRA-MoCo v3. We integrate our Stain Reconstruction Augmentation (SRA) as well as additional contrastive loss terms (CL3 and CL4) into MoCo v3. }
    \label{fig:pathmoco_flowchart}
    }}
\end{figure*}

\section{Related works}
\subsection{Pathology-specific augmentation}
In the context of pathology-specific augmentations, various methods have been proposed to address domain-specific challenges and variations in histopathological images. Shen et al. \cite{Shen2022-fk-RandstainNA} introduced RandStainNA, which generates random template slides for color normalization and augmentation in HSV, LAB, and HED color spaces to tackle variations in staining and colors across different slides and datasets. Additionally, Gullapally et al. \cite{Gullapally2023-rt-SDOTA} addressed inter-laboratory and scanner variability through Scanner Transform (ST) and Stain Vector Augmentation (SVA), enhancing out-of-distribution performance on tasks such as tissue segmentation.

A fundamental operation for many pathology-specific augmentations is stain separation, which isolates single-channel images in Optical Density (OD) space based on the Beer-Lambert law \cite{Lambert1760-xs, Beer1852-dl}. Augmentations can then be applied independently to each stain channel. In \cite{Yang2022-er-cs-co}, perturbations are applied to the stain separation matrix to deal with the errors in separation matrix calculation. In \cite{tellez2018, Tellez2018-wj}, each channel is randomly scaled and biased within a narrow range before converting back to RGB space. However, the maximum possible intensity after augmentation is still influenced by the original image's maximum intensity. \cite{Chang2021-th-mixup} also utilizes this method, along with random stain matrix interpolation, to handle domain variations across datasets by incorporating information from both source and target data.

\subsection{Self-supervised learning and contrastive loss}
Recent advancements in self-supervised learning \cite{Zbontar2021-si-BT, Grill2020-wh-BYOL} have introduced novel frameworks for learning robust and accurate features across various datasets. Barlow Twins \cite{Zbontar2021-si-BT} encourages two augmented views of the same input to produce similar but decorrelated representations by minimizing the cross-correlation between them. DINO \cite{Caron2021-vx-dino, Oquab2023-ja-dinov2p1, Darcet2023-mg-dinov2p2} employs a student-teacher framework within a vision transformer architecture, without requiring labeled data. In DINO, the student encoder attempts to mimic the teacher encoder, which is updated based on an exponential moving average (EMA). Unlike the student encoder, which processes both global and local views of the images, the teacher encoder only receives global views. Building on the DINO framework, PathDino \cite{Alfasly2023-wy-pathdino} combines lightweight transformers with a novel 360° rotation augmentation (HistoRotate), achieving robust performance across 12 diverse pathology datasets.

Contrastive learning is one of the most widely used and fundamental approaches in self-supervised learning pipelines. For instance, the SimCLR framework \cite{Chen2020-ta-simclr, Chen2020-pf-simclr2} utilizes NT-Xent loss on strongly augmented views of images, aiming to minimize the distance between different views of the same image while maximizing the distance between views of different images. In contrast, SwAV \cite{Caron2020-mj-SwAV} employs a cluster-based contrastive learning approach rather than a pairwise one, using a swapped prediction mechanism to encourage the features of the same cluster to be as invariant as possible. The MoCo framework \cite{He2019-jc-mocov1, Chen2020-mz-mocov2, Chen2021-gz-mocov3} takes a distinct approach to contrastive learning called momentum contrast. In this framework, there are two encoders: a momentum encoder and a query encoder. The momentum encoder is updated as the exponential moving average of the query encoder and maintains a consistent dictionary of negative samples, enabling contrastive learning with a large and diverse set of negatives across both current and previous batches. Other popular contrastive learning methods include iBOT \cite{Zhou2021-oa-iBOT}, RePre \cite{Wang2022-im-RePre} and RECON \cite{Qi2023-el-recon}.

\section{Methods}
\subsection{An overview of SRA-MoCo v3}
\begin{figure}[ht]
    \centering
    \includegraphics[width=1.0\linewidth]{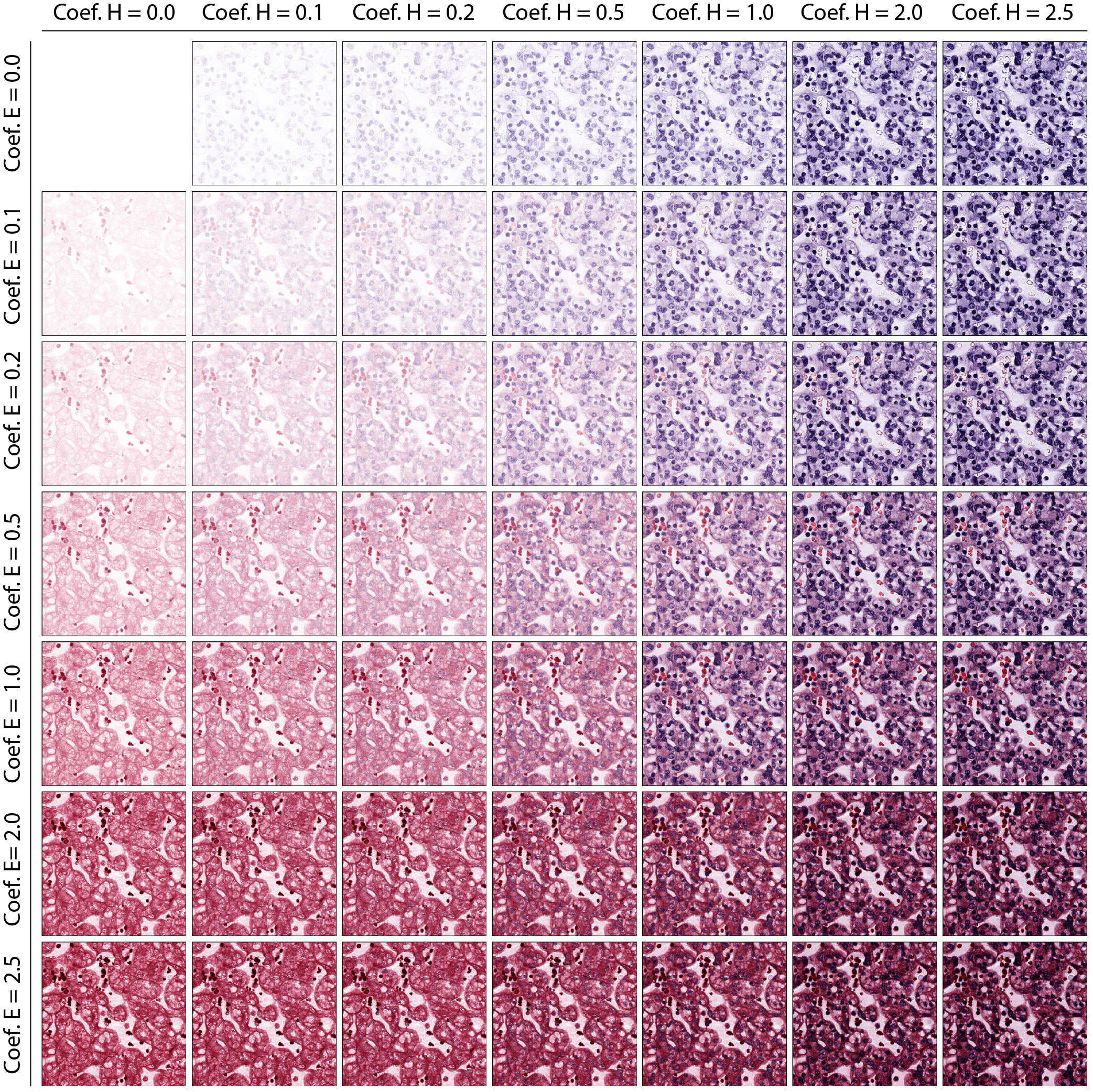}
    \caption{Examples of augmentations by SRA with different target strengths of H channel and E channel. }
    \label{fig:augm_examples}
\end{figure}

\cref{fig:pathmoco_flowchart} shows the overall workflow of SRA-MoCo v3. First, all pixels within the tissue regions of an H\&E whole slide image (WSI) are collected to analyze the max intensity (strength) of each stain in the current WSI, where the intensity is measured on Optical Density (OD) space after stain separation process. 
To perform stain reconstruction augmentation, we predefine an absolute range for the target strength of each stain and map the real strength of each stain to a random value in this target range. Unlike the approach in \cite{tellez2018,Tellez2018-wj,Chang2021-th-mixup}, which only slightly adjusts the strength of each stain within a relative range by multiplying a random factor between 0.95 and 1.05, our SRA directly defines a much broader absolute range for target strength of each stain channel. For instance, if the target range is set between 0.5 and 2, and the original strength of a particular stain in a WSI is 2, this indicates a deeply stained image. Augmenting this stain channel only makes the new maximum intensity fall between 0.5 and 2, without surpassing the original strength. While the traditional augmentation makes the new maximum intensity fall between 1.9 and 2.1. Our method allows for more extensive and stronger augmentations while ensuring the strength remains within an appropriate range. Moreover, inspired by multi-modal contrastive learning \cite{Chai2022-gp-multimodal,Zhang2022-ws-cotraining}, we define specific probabilities for excluding one stain channel, allowing only the other stain channel to remain after augmentation. Examples of images resulting from our stain reconstruction augmentations, including single stain channel images, can be seen in \cref{fig:augm_examples}.

\begin{figure*}[ht]
    \centering
    \includegraphics[width=1.0\linewidth]{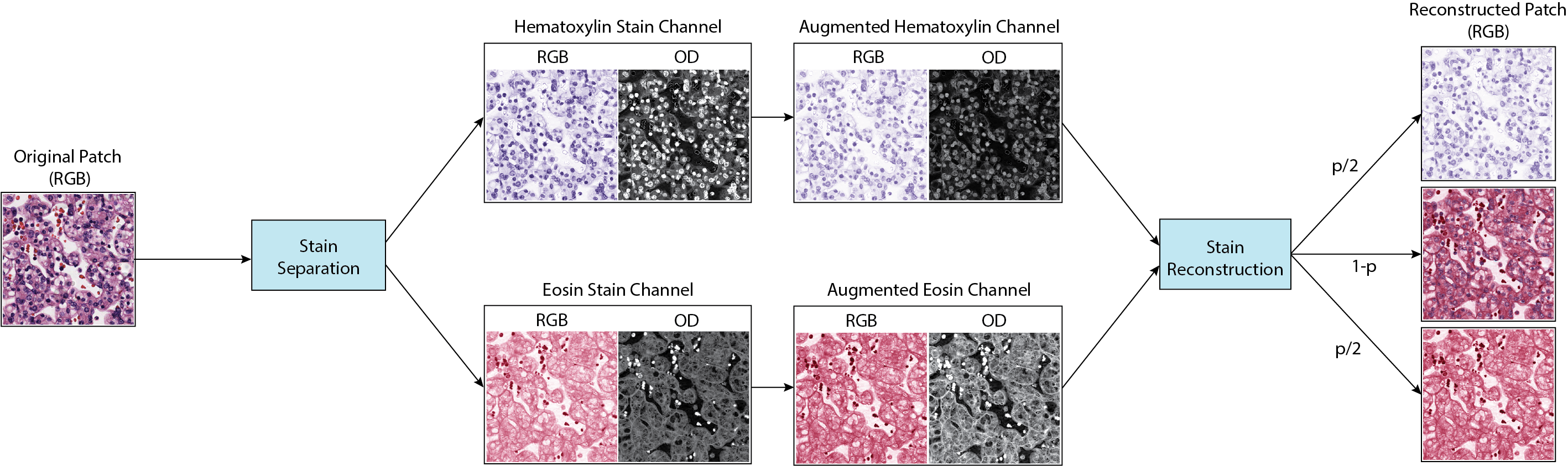 }
    \caption{Demonstration of stain reconstruction augmentation (SRA). Single stain images are shown in both RGB space and OD space. The augmentations are performed on each stain channel independently. There is a probability of p that only single channel is adopted.}
    \label{fig:SRA}
\end{figure*}

After stain reconstruction augmentation, additional general image augmentation methods are applied to introduce further variations. MoCo v3 is used as the backbone for histopathology image representation learning. In MoCo v3, the contrastive loss is computed between queries from the query encoder and keys from the momentum encoder using different augmentations. However, there are no loss terms that specifically focus on contrastive learning between different augmentations from the same encoder, such as the contrastive loss between identical queries with different augmentations. Given the substantial variations introduced by stain reconstruction augmentation, we further explored the addition of contrastive loss terms that focus solely on augmentations. More details are provided in the following subsections.

\subsection{Stain reconstruction augmentation}

As shown in \cref{fig:SRA}, for an H\&E image, we first perform stain separation using the algorithm from \cite{Macenko2009-su} to obtain single-stain images. For each whole slide image (WSI), the RGB pixel values are mapped into Optical Density (OD) space $(OD_R, OD_G, OD_B)$ according to the Beer-Lambert law \cite{Beer1852-dl,Lambert1760-xs}, where higher OD values indicate stronger stains. All pixels are mapped to the same OD space. Based on the distribution of these pixels in OD space, three unit vectors, $V_H$, $V_E$, and $V_{Residual}$, are derived, which allow for the decomposition of OD values as shown below:
\begin{equation}
  (OD_R, OD_G, OD_B)=\alpha V_H + \beta V_E + \gamma V_{Residual}
  \label{eq:tsa_od}
\end{equation}

In each slide, we calculate the values of $\alpha$ and $\beta$ for each tissue pixel, where $\alpha$ represents the proportion of Hematoxylin stain and $\beta$ represents the proportion of Eosin stain for each pixel. For each WSI, we define $H_{max}$ (maximum intensity of Hematoxylin stain) as the 99th percentile of all $\alpha$ values in the tissue pixels of the slide and similarly define $E_{max}$ (maximum intensity of Eosin stain) as the 99th percentile of all $\beta$ values. This slide-level operation is performed because the tissues in the same slide are stained and stored under identical conditions. Additionally, due to the large number of pixels in a slide, slide-level operation provides stable results.

In the first step of implementing our stain reconstruction augmentation, we predefine global target ranges for new $H_{max}$ and $E_{max}$ after augmentation. For each training image, following the stain separation process, we independently and randomly select coefficients $coef_H$ and $coef_E$ from within the target ranges. We then multiply $\alpha$ by $coef_H/H_{max}$ and $\beta$ by $coef_E/E_{max}$ to randomly adjust the stain strength. Furthermore, we introduce a hyperparameter $p$, which defines the probability of randomly setting either $coef_H$ or $coef_E$ to zero during stain reconstruction augmentation, thereby creating additional variations.
Finally, after all processes are complete, we reconstruct the images back into RGB space from OD space, based on the new proportions of Hematoxylin and Eosin stains. The code for stain reconstruction augmentation can be found in Supplementary material.

\begin{figure*}[ht]
    \centering
    \includegraphics[width=1.0\linewidth]{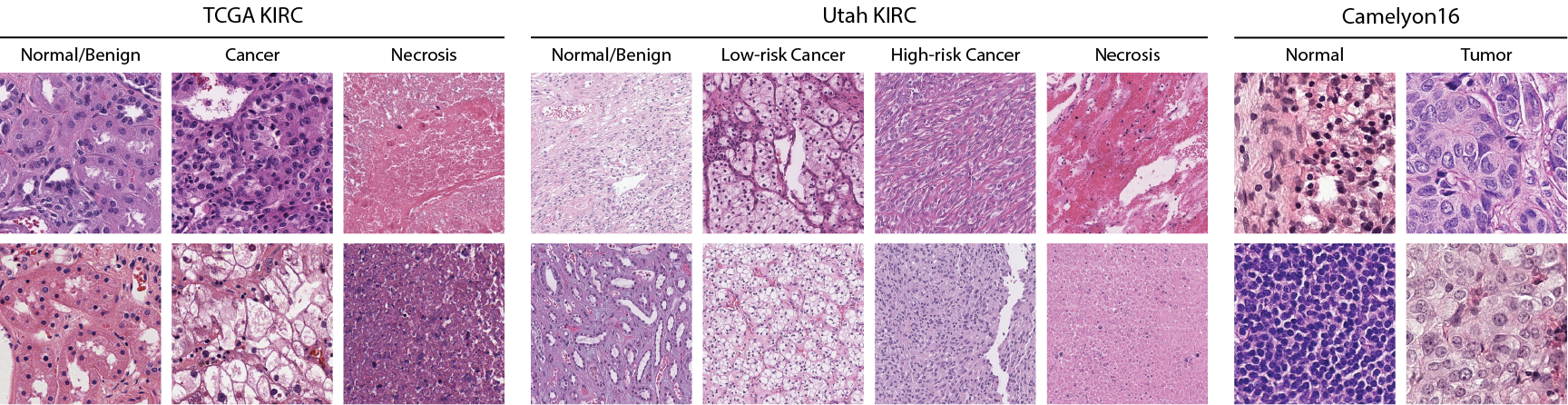}
    \caption{Patch examples from different classes and different datasets.}
    \label{fig:datasets_patch_examples}
\end{figure*}

\subsection{Contrastive learning loss}
We adopted MoCo v3 \cite{Chen2021-gz-mocov3} as the platform for our contrastive learning framework. MoCo v3 consists of two encoders: a momentum encoder and a query encoder. The model aims to match an encoded query $q$ from the query encoder to a dictionary of encoded keys from the momentum encoder.
The contrastive loss term between queries and keys can be written as:
\begin{equation}
  CL(k, q)=-\log \left( \frac{\exp \left( q \cdot k^+ / \tau \right)}{\exp \left( q \cdot k^+ / \tau \right) + \sum_{k^-} \exp \left( q \cdot k^- / \tau \right)} \right)
  \label{eq:cl_form}
\end{equation}

where $\tau$ is temperature hyperparameter. As shown in \cref{fig:pathmoco_flowchart}, each input image is augmented twice, and the augmented images are then passed through both the query encoder and the momentum encoder. Let $f_M^{Aug1}$ represent the output features from the momentum encoder after the first augmentation, $f_M^{Aug2}$ the output after the second augmentation, $f_B^{Aug1}$ the output from the query encoder after the first augmentation, and $f_B^{Aug2}$ the output after the second augmentation. The original contrastive loss can be written as:

\begin{equation}
  CL_{ori}=CL_1(f_M^{Aug1}, f_B^{Aug2}) + CL_2(f_M^{Aug2}, f_B^{Aug1})
  \label{eq:cl_ori}
\end{equation}

In the standard contrastive loss pairs, both the augmentations and the encoders differ. To make the contrastive learning process more sensitive to augmentations, especially given the introduction of strong stain augmentation, we introduce the following additional loss terms:

\begin{equation}
  CL_{aug}=CL_3(f_B^{Aug1}, f_B^{Aug2}) + CL_4(f_M^{Aug1}, f_M^{Aug2})
  \label{eq:cl_aug}
\end{equation}

Thus, the overall contrastive learning loss becomes $CL_{ori} + CL_{aug}$. We will evaluate the benefits of incorporating these extra loss terms through ablation studies in the Experiments section.

\begin{table}[t]
\centering
{\small{
\setlength{\tabcolsep}{7.5pt}  
\begin{tabular}{ccccc}
\toprule
Set                                  & \makecell{Normal/\\Benign}              & Cancer   & Necrosis    & Total                                    \\ \midrule
\multirow{2}{*}{\vspace{-8pt}Train}  & \multicolumn{3}{c}{1,373,684 Unlabeled} &                          \multirow{2}{*}{\vspace{-8pt}1,646,665}  \\ \cmidrule(l){2-4}
                                     & 84,578                                  & 180,471  & 7,932       &                                          \\ \midrule
Val.                                 & 19,638                                  & 79,382   & 1,301       & 100,321                                  \\ \midrule
Test                                 & 15,323                                  & 62,565   & 6,168       & 84,056                                   \\ \bottomrule
\end{tabular}
\caption{Summary of the number of patches for each category in each set on TCGA KIRC dataset. The training set includes both labeled and unlabeled patches.}
\label{tab:TCGA_patch_count}
}}
\end{table}

\section{Experiments}
\subsection{Datasets}
We conducted all our experiments on three datasets: The Cancer Genome Atlas Kidney Renal Clear Cell Carcinoma (TCGA KIRC) dataset \cite{gdc-portal-ju}, the Utah KIRC dataset \cite{Zhang2023-lo-class-m}, and the Camelyon 16 dataset \cite{Ehteshami-Bejnordi2017-ai-camelyon16}, all of which consist of H\&E-stained images. The first two datasets contain kidney WSIs, while the Camelyon 16 dataset consists of breast WSIs. The labels for the TCGA KIRC and Utah KIRC images were obtained from \cite{Zhang2023-lo-class-m}, and the labels for the Camelyon 16 dataset were sourced from \cite{camelyon16-yf-website}. Examples of patches from the three datasets are shown in \cref{fig:datasets_patch_examples}.

From the TCGA KIRC dataset, we collected 420 WSIs in total, with 300 slides used for training, 60 slides for validation, and 60 slides for testing. The dataset provides 1,646,665 tissue patches of size 400x400 at 20X resolution from the 300 training slides for self-supervised contrastive learning, as well as labeled patches for 3-class classification tasks. More details can be found in \cref{tab:TCGA_patch_count}.

In the Utah KIRC dataset, there are 49 slides from different patients, with 32 slides for training, 10 slides for validation, and 7 slides for testing. The 32 training slides provide 208,291 tissue patches of size 400x400 at 10X resolution for self-supervised contrastive learning. This dataset also includes labeled patches for 4-class classification tasks. For more details, please refer to \cref{tab:Utah_patch_count}.
\begin{table}[t]
\centering
{\small{
\setlength{\tabcolsep}{3.3pt} 
\begin{tabular}{cccccc} 
\toprule
Set                       & \makecell{Normal/\\Benign}            & \makecell{Low-risk \\ Cancer} & \makecell{High-risk \\ Cancer}  & Necrosis  & Total                                   \\ \midrule
\multirow{2}{*}{\vspace{-8pt}Train}    & \multicolumn{4}{c}{171,113 Unlabeled} &                                                                  \multirow{2}{*}{\vspace{-8pt}208,291}   \\ \cmidrule(l){2-5}
                          & 28,497                                & 2,044                         & 2,522                           & 4,115     &                                         \\ \midrule
Val.                      & 5,472                                 & 416                           & 334                             & 2,495     & 8,117                                   \\ \midrule
Test                      & 7,263                                 & 598                           & 389                             & 924       & 9,174                                   \\ \bottomrule
\end{tabular}
\caption{Summary of the number of patches for each category in each set on Utah KIRC dataset. The training set includes both labeled and unlabeled patches.}
\label{tab:Utah_patch_count}
}}
\end{table}

\begin{figure*}[ht]
    \centering
    \includegraphics[width=1.0\linewidth]{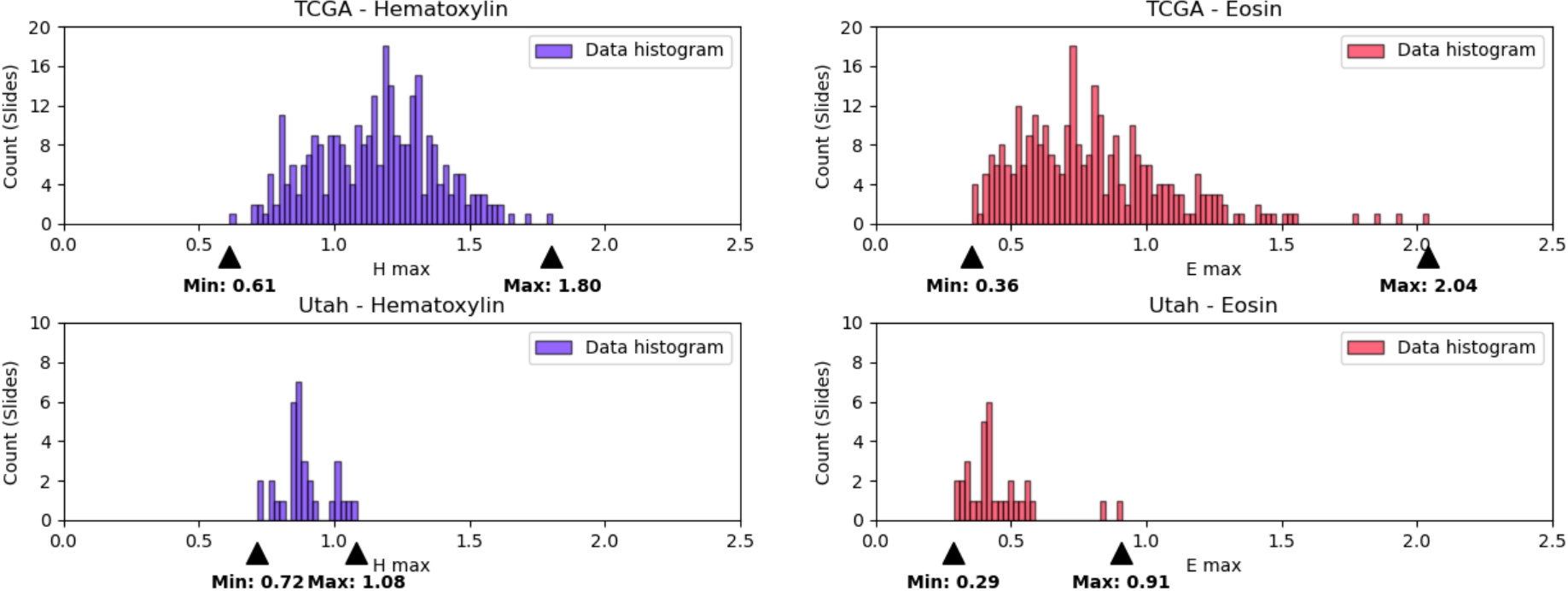}
    \caption{Distributions of strengths of Hematoxylin stain and Eosin stain in Optical Density (OD) space on TCGA training set and Utah training set.}
    \label{fig:hists}
\end{figure*}

Camelyon 16 is another public dataset consisting of breast cancer slides for multi-instance learning. Instead of patch-level labels, it provides only slide-level labels for binary classification between Normal and Tumor. The training set contains 160 Normal slides and 111 Tumor slides, while the test set includes 80 Normal slides and 50 Tumor slides. We randomly selected $10\%$ of the training slides to create a validation set. We followed the standard procedure as other papers \cite{Zhang2022-xb-dtfdmil} to crop patches in 20X resolution. 

\subsection{Experiment settings}
In our experiments, we began by pre-training the encoders using self-supervised contrastive feature representation learning, followed by testing the pre-trained encoders on downstream tasks. The first scenario involves performing both pre-training and downstream classification tasks on the same dataset. The second scenario involves pre-training the encoders on one dataset and evaluating them on downstream tasks from a different dataset. We implemented the first scenario on the TCGA KIRC and Utah KIRC datasets separately. For the second scenario, we pre-trained the encoders on the TCGA KIRC dataset, then evaluated them on the Utah KIRC and Camelyon 16 separately. Since the patches in the Utah dataset have a resolution of 10X, compared to 20X in the TCGA KIRC dataset, we also cropped patches at 10X resolution from the TCGA KIRC training slides to ensure resolution consistency between pre-training and downstream tasks.

For stain reconstruction augmentation, we first predefined a range for the target strengths of the Hematoxylin and Eosin stains. We calculated the distributions of $H_{max}$ and $E_{max}$ across all training slides in both the TCGA KIRC and Utah KIRC datasets. As shown in \cref{fig:hists}, the stain strengths ($H_{max}$ and $E_{max}$) varies across different slides in both datasets. In our experiments, we tested two sets of target ranges. The first set marginally covers the distribution of stain intensities in TCGA KIRC, with a target range of [0.5, 2.0] for new $H_{max}$ and [0.2, 2.0] for new $E_{max}$. The second set has wider target ranges to introduce stronger augmentations, with both new $H_{max}$ and new $E_{max}$ set to [0.1, 2.5]. We selected this range because further widening would cause saturation when reconstructing images from Optical Density (OD) space back to RGB space.

After applying stain reconstruction augmentation, we followed the standard MoCo v3 training procedure, including additional general augmentations, with ResNet50 \cite{He2015-zb-resnet} as backbone. The encoders consistently received 224x224 patches as inputs except for PathDino, where 512x512 patches were used after image resizing. We also included an option in our code to add the augmentation contrastive loss, $CL_{aug}$. Moreover, we tracked the contrastive loss on the validation set and selected the training epoch with the lowest contrastive loss on the validation set. In the downstream tasks on the TCGA KIRC and Utah KIRC, we used a balanced sampler for training and balanced accuracy for evaluation, as the patch labels were highly imbalanced. For the downstream tasks on the Camelyon 16 dataset, we employed DTFD-MIL \cite{Zhang2022-xb-dtfdmil}, as it has been shown to be one of the best multi-instance learning models for Camelyon 16 and other datasets. Stain reconstruction augmentation was only applied during pre-training, not in any downstream tasks, where the MoCo v3's image classification code was used.

All experiments were conducted using Python 3.11.4, PyTorch 2.0.1, torchvision 0.15.2, and CUDA 11.8 on NVIDIA RTX A6000 GPUs. With a batch size of 512, the MoCo v3 pre-training required a total of around 63 GB of memory and approximately 20 hours of runtime across two parallel GPUs.

\subsection{Pre-training and downstream task on same datasets}
We first evaluated SRA-MoCo v3 by performing contrastive feature representation learning and downstream classification on the same datasets (TCGA KIRC or Utah KIRC). Since there is no domain shift in this scenario, we consistently used the first option’s target range for stain strengths, which has narrower limits (H: [0.5, 2.0], E: [0.2, 2.0]), and set the probability of generating pure Hematoxylin or pure Eosin images to zero. After pre-training, we froze the pre-trained encoders and performed classification on the same dataset. For reference, we also compared the classification results with other state-of-the-art foundation encoder models, including PathDino pre-trained on 11,765 TCGA slides by \cite{Alfasly2023-wy-pathdino}, as well as Barlow Twins, SwAV, and MoCo v2 pre-trained on 20,994 TCGA slides and 15672 from the TULIP dataset by \cite{Kang2023-oo-benchmarking}. Additionally, we included the basic ResNet50 model in a fully supervised learning setting for comparison.

As shown in \cref{tab:tcga_on_tcga} and \cref{tab:utah_on_utah}, our SRA-MoCo v3 significantly outperforms standard MoCo v3 and stain augmentations from \cite{tellez2018, Tellez2018-wj}. On the TCGA KIRC dataset, SRA-MoCo v3 surpasses MoCo v3 by 4.25\% (83.62\% vs. 79.37\%) in balanced accuracy, and SRA also outperforms the traditional stain augmentation (TSA) \cite{tellez2018, Tellez2018-wj} by 2.12\% (83.62\% vs. 81.50\%). On the Utah KIRC dataset, SRA-MoCo v3 exceeds MoCo v3 by 2.08\% (95.85\% vs. 93.77\%), and MoCo v3 plus TSA by 1.85\% (95.85\% vs. 94.00\%). The comparison with other foundation models further demonstrates the advantages of SRA-MoCo v3. Despite these state-of-the-art models being pre-trained on much larger datasets, SRA-MoCo v3 still reaches the best.

\begin{table}[t]
\centering
{\small{
\setlength{\tabcolsep}{1pt}  
\renewcommand{\arraystretch}{1.2}
\begin{tabular}{ccc}
\toprule
\makecell{Pre-trained\\Dataset}                                    & \makecell{Model}  & \makecell{Balanced Accuracy\\(TCGA KIRC)}   \\ \midrule
ImageNet                                                           & ResNet50          & 69.97 \( \pm \) 5.59                   \\ \midrule 
\multirow{3}{*}{\makecell{TCGA + TULIP\\32.6M patches}} & Barlow Twins                & 81.59 \( \pm \) 2.65                   \\ 
                                                                   & MoCo v2           & 79.04 \( \pm \) 0.11                   \\ 
                                                                   & SwAV              & 77.43 \( \pm \) 1.15                   \\ \midrule
\makecell{11,765 TCGA Slides\\6.1M patches}                              & PathDino          & 76.92 \( \pm \) 6.22                   \\ \midrule
\multirow{3}{*}{\makecell{TCGA KIRC\\300 Slides\\1.6M patches}}    & MoCo v3           & 79.37 \( \pm \) 1.18                   \\ 
                                                                   & MoCo v3 + TSA     & 81.50 \( \pm \) 0.23                   \\ 
                                                                   & SRA-MoCo v3          & 83.62 \( \pm \) 0.28                   \\ \bottomrule
\end{tabular}
\caption{Performance of pre-trained models on TCGA KIRC dataset compared to foundation models (20X magnification). TSA represents traditional stain augmentation methods \cite{tellez2018,Tellez2018-wj}}.
\label{tab:tcga_on_tcga}
}}
\end{table}

\subsection{Pre-training and downstream task on different datasets}
Although our SRA-MoCo v3 demonstrates outstanding results as discussed in the previous subsection, the more common application of self-supervised learning is to pre-train an encoder on large datasets and then apply the frozen encoder to other datasets. To evaluate SRA-MoCo v3 in a more practical setting, we pre-trained SRA-MoCo v3 on 300 training slides from the TCGA KIRC dataset and subsequently evaluated the encoders on the Utah KIRC dataset and the Camelyon 16 dataset separately.

Considering the presence of domain shift, we used the second option's wider target range for stain strength in the stain reconstruction augmentation (H: [0.1, 2.5], E: [0.1, 2.5]) and set the probability of generating pure Hematoxylin or pure Eosin images to 0.1. We also compared the classification results with other state-of-the-art foundation models pre-trained on large datasets, including PathDino, Barlow Twins, SwAV, and MoCo v2. For the downstream multi-instance learning on the Camelyon 16 dataset, we also used the ImageNet pre-trained ResNet50 as the encoder within the DTFD-MIL framework as a baseline.

Based on \cref{tab:transfer_learning}, on the Utah KIRC dataset, we observed 2.8\% improvement with SRA-MoCo v3 compared to MoCo v3. Both SRA-MoCo v3 and MoCo v3 pre-trained on the TCGA KIRC dataset outperform those pre-trained on the Utah KIRC dataset itself. The most likely reason is that the TCGA KIRC dataset contains significantly more patches. Even though domain shift is present, both datasets focus on kidney cancer, which mitigates the impact. When comparing the results with foundation models, it becomes evident that pre-training and downstream classification both on kidney images offer distinct advantages. 

\begin{table}[t]
\centering
{\small{
\setlength{\tabcolsep}{1pt} 
\renewcommand{\arraystretch}{1.2} 
\begin{tabular}{ccc}
\toprule
\makecell{Pre-trained\\Dataset}                                     & \makecell{Model}  & \makecell{Balanced Accuracy\\(Utah KIRC)}    \\ \midrule
ImageNet                                                            & ResNet50          & 87.76 \( \pm \) 0.10                    \\ \midrule 
\multirow{3}{*}{\makecell{TCGA + TULIP\\32.6M patches}}  & Barlow Twins                & 90.23 \( \pm \) 1.81                    \\ 
                                                                    & MoCo v2           & 91.45 \( \pm \) 0.44                    \\ 
                                                                    & SwAV              & 94.96 \( \pm \) 1.04                    \\ \midrule
\makecell{11,765 TCGA Slides\\6.1M patches}                              & PathDino          & 92.14 \( \pm \) 1.65                    \\ \midrule
\multirow{3}{*}{\makecell{Utah KIRC\\49 Slides\\0.2M patches}}      & MoCo v3           & 93.77 \( \pm \) 0.86                    \\ 
                                                                    & MoCo v3 + TSA     & 94.00 \( \pm \) 0.26                    \\ 
                                                                    & SRA-MoCo v3          & 95.85 \( \pm \) 0.34                    \\ \bottomrule
\end{tabular}
\caption{Performance of pre-trained models on Utah KIRC dataset compared to foundation models (10X magnification). TSA represents traditional stain augmentation methods \cite{tellez2018, Tellez2018-wj}.}
\label{tab:utah_on_utah}
}}
\end{table}

\begin{table*}[ht]
\centering
{\small{
\setlength{\tabcolsep}{8.5pt} 
\begin{tabular}{cc|c|ccc} 
\toprule
\makecell{Pre-trained\\Dataset}                                    & \makecell{Model}  & \makecell{Balanced Accuracy\\Utah KIRC} & \makecell{F1-score\\Camelyon16}  & \makecell{Accuracy\\Camelyon16}  & \makecell{Balanced Accuracy\\Camelyon16}  \\ \hline
ImageNet                                                           & ResNet50          & 87.76 \( \pm \) 0.10               & 0.8372 \( \pm \) 0.0149          & 88.37 \( \pm \) 0.78             & 86.54 \( \pm \) 1.35                      \\ \hline 
\multirow{3}{*}{\makecell{TCGA + TULIP\\32.6M patches}} & Barlow Twins                & 90.23 \( \pm \) 1.81               & 0.9019 \( \pm \) 0.0100          & 93.02 \( \pm \) 0.78             & 91.35 \( \pm \) 0.67                      \\ 
                                                                   & MoCo v2           & 91.45 \( \pm \) 0.44               & 0.9291 \( \pm \) 0.0049          & 94.83 \( \pm \) 0.45             & 93.73 \( \pm \) 0.13                      \\ 
                                                                   & SwAV              & 94.96 \( \pm \) 1.04               & 0.9264 \( \pm \) 0.0098          & 94.57 \( \pm \) 0.78             & 93.65 \( \pm \) 0.63                      \\ \hline
\makecell{11,765 TCGA Slides\\6.1M patches}                              & PathDino          & 92.14 \( \pm \) 1.65               & 0.9176 \( \pm \) 0.0252          & 93.80 \( \pm \) 2.05             & 93.15 \( \pm \) 1.74                      \\ \hline
\multirow{3}{*}{\makecell{TCGA KIRC\\300 Slides\\0.4M/1.6M patches}}    & MoCo v3           & 95.32 \( \pm \) 0.30               & 0.8075 \( \pm \) 0.0135          & 85.79 \( \pm \) 1.79             & 84.33 \( \pm \) 0.89                      \\ 
                                                                   & MoCo v3 + TSA     & 94.17 \( \pm \) 0.82               & 0.8268 \( \pm \) 0.0163          & 87.60 \( \pm \) 2.05             & 85.65 \( \pm \) 0.94                      \\ 
                                                                   & SRA-MoCo v3          & 98.12 \( \pm \) 0.15               & 0.9207 \( \pm \) 0.0084          & 94.31 \( \pm \) 0.44             & 92.91 \( \pm \) 0.95                      \\ \bottomrule
\end{tabular}

\caption{Performance of pre-trained models on Utah KIRC dataset and Camelyon16 dataset, compared to foundation models. TSA represents traditional stain augmentation methods \cite{tellez2018, Tellez2018-wj}. In 300 slides from TCGA KIRC, 363,225 10X patches are used 
 in pre-training for downstream task on Utah KIRC, 1,646,665 20X patches are used in pre-training for downstream task on Camelyon 16.}
\label{tab:transfer_learning}
}}
\end{table*}

\begin{table*}[ht]
\centering
{\small{
\setlength{\tabcolsep}{7.5pt} 
\begin{tabular}{cccc|c|ccccc}
\toprule
 \makecell{Range\\$coef_H$} & \makecell{Range\\$coef_E$} & \makecell{\textit{p}(only\\H or E)} &  \makecell{Extra\\Loss} & \makecell{Balanced Acc.\\Utah KIRC} &\makecell{F1-score\\Camelyon16} & \makecell{Accuracy\\Camelyon16} & \makecell{Balanced Acc.\\Camelyon16} \\ \hline
  N/A                & N/A                 & 0                       & \( - \)                 & 95.32 \( \pm \) 0.30           & 0.8075 \( \pm \) 0.0135        & 85.79 \( \pm \) 1.79\           & 84.33 \( \pm \) 0.89\  \\
  N/A                & N/A                 & 0                       & \( CL_{aug} \)          & 95.34 \( \pm \) 0.41           & 0.8457 \( \pm \) 0.0253        & 88.11 \( \pm \) 1.19\           & 87.78 \( \pm \) 2.44\  \\
  {[}0.2, 2.0{]}     & {[}0.5, 2.0{]}      & 0                       & \( - \)                 & 96.51 \( \pm \) 0.37           & 0.8244 \( \pm \) 0.0047        & 87.34 \( \pm \) 0.45\           & 85.58 \( \pm \) 0.42\  \\
  {[}0.1, 2.5{]}     & {[}0.1, 2.5{]}      & 0                       & \( - \)                 & 96.95 \( \pm \) 0.76           & 0.8494 \( \pm \) 0.0154        & 89.67 \( \pm \) 0.89\           & 87.19 \( \pm \) 1.25\  \\
  {[}0.2, 2.0{]}     & {[}0.5, 2.0{]}      & 0                       & \( CL_{aug} \)          & 96.86 \( \pm \) 0.17           & 0.8341 \( \pm \) 0.0060        & 88.37 \( \pm \) 0.78\           & 86.15 \( \pm \) 0.33\  \\
  {[}0.1, 2.5{]}     & {[}0.1, 2.5{]}      & 0                       & \( CL_{aug} \)          & 98.09 \( \pm \) 0.12           & 0.8596 \( \pm \) 0.0213        & 89.15 \( \pm \) 2.05\           & 88.75 \( \pm \) 1.40\  \\
  {[}0.2, 2.0{]}     & {[}0.5, 2.0{]}      & 10\%                    & \( CL_{aug} \)          & 97.41 \( \pm \) 0.08           & 0.9079 \( \pm \) 0.0150        & 93.28 \( \pm \) 1.18\           & 92.08 \( \pm \) 1.02\  \\
  {[}0.1, 2.5{]}     & {[}0.1, 2.5{]}      & 10\%                    & \( CL_{aug} \)          & 98.12 \( \pm \) 0.15           & 0.9207 \( \pm \) 0.0084        & 94.31 \( \pm \) 0.44\           & 92.91 \( \pm \) 0.95\  \\ \bottomrule
\end{tabular}
\caption{Ablation study results showing the impact of each component of SRA-MoCo v3 on Utah KIRC dataset and Camelyon 16 dataset.}
\label{tab:ablation}
}}
\end{table*}

The multi-instance learning results on the Camelyon 16 dataset can also be found in \cref{tab:transfer_learning}. The MoCo v3 encoder pre-trained on the TCGA KIRC dataset shows lower performance compared to encoders pre-trained on ImageNet. One possible explanation is the substantial domain shift between kidney cancer and breast cancer slides. This is further illustrated in \cref{fig:datasets_patch_examples}, which highlights the significant differences between TCGA kidney cancer patches and Camelyon 16 breast cancer patches. However, with our stain reconstruction augmentation and our additional augmentation contrastive loss, SRA-MoCo v3 improves the balanced accuracy by 8.53\% compared to MoCo v3, while MoCo v3 + TSA exceeds MoCo v3 by only 1.82\%. When comparing results with other foundation models pre-trained on much larger datasets with a greater variety of organ types, SRA-MoCo v3 surpasses Barlow Twins and achieves comparable F1 score, accuracy, and balanced accuracy to PathDino, SwAV, and MoCo v2. Our pre-trained SRA-MoCo v3 encoder will be released upon publication.

\subsection{Ablation studies}
We also conducted ablation studies to carefully analyze the impact of each component in SRA-MoCo v3.

The ablation studies for transfer learning from the TCGA KIRC dataset to the Utah KIRC dataset are presented in \cref{tab:ablation}. From the results, we found that simply adding the augmentation contrastive loss to MoCo v3 does not yield any improvement. However, this loss becomes effective when combined with stain reconstruction augmentation. Using a wider target range for stain strength and incorporating the possibility of generating pure Hematoxylin or pure Eosin images in stain reconstruction augmentation also proved beneficial. During transfer learning, the slides exhibit significantly more variation across datasets, and SRA-MoCo v3 addresses this by generating highly augmented images that are also clinically meaningful.

Lastly, we evaluated the contribution of each component in SRA-MoCo v3 through downstream tasks on the Camelyon 16 dataset. We observed that both the augmentation contrastive loss and stain reconstruction augmentation independently improve performance. When combining the augmentation contrastive loss with stain reconstruction augmentation using a wider target range of stain variation, we achieved a 4.43\% improvement in balanced accuracy. Unlike transfer learning from the TCGA KIRC to the Utah KIRC dataset, where setting the probability of converting augmented images to pure Hematoxylin or pure Eosin to 0.1 resulted in a minor boost, we observed a 4.16\% increase in balanced accuracy on Camelyon 16 by simply adjusting this probability from 0 to 0.1, which implies that stronger augmentations are more beneficial on cases with stronger domain shift. All the adjustments contributed to SRA-MoCo v3 outperforming MoCo v3 by 8.58\%.

\section{Conclusion}
In this paper, we propose a novel framework for contrastive representation learning in pathology images. Specifically, we introduce a new stain reconstruction augmentation method and an augmentation contrastive loss. Experiments on downstream tasks, including image classification and multi-instance learning, demonstrate significant improvements when applying our algorithms to MoCo v3. Despite being pre-trained on much smaller datasets, SRA-MoCo v3 achieves comparable or superior performance in various self-supervised learning and downstream tasks compared to other state-of-the-art foundational encoders.

\section*{Acknowledgements}
The work was funded in part by NIH 1R21CA277381. The Computational Oncology Research Initiative (CORI) at the Huntsman Cancer Institute, The Department of Pathology at the University of Utah and ARUP Laboratories had also supported the work.
Part of the results presented in this paper are based upon data generated by the TCGA Research Network: \hyperlink{https://www.cancer.gov/tcga}{https://www.cancer.gov/tcga}

{\small
\bibliographystyle{ieee_fullname}
\bibliography{bibliography.bib}
}
\end{document}